\documentclass[10pt,twocolumn,letterpaper]{article}

\usepackage[pagenumbers]{cvpr} 

\usepackage{pifont}
\usepackage{algorithm}
\usepackage{algorithmic}
\usepackage[table]{xcolor}

\definecolor{cvprblue}{rgb}{0.21,0.49,0.74}
\usepackage[pagebackref,breaklinks,colorlinks,allcolors=cvprblue]{hyperref}

\def\paperID{} 
\def\confName{CVPR}
\def\confYear{2026}

\title{Masked Representation Modeling for Domain-Adaptive Segmentation}

\author{
	Wenlve Zhou$^{1}$ \quad
	Zhiheng Zhou$^{1}$\thanks{Corresponding author: zhouzh@scut.edu.cn} \quad
	Tiantao Xian$^{1}$ \quad
	Yikui Zhai$^{2}$  \quad
	Weibin Wu$^{3}$  \quad
	Biyun MA$^{1}$ \\ 
	\\
	$^{1}$South China University of Technology \\
	$^{2}$School of Electronic and Information Engineering, Wuyi University \\
	$^{3}$South China Agricultural University \\
}

\begin{document}
\maketitle

\begingroup
\renewcommand\thefootnote{}
\footnotetext{Code available at \url{https://github.com/Wenlve-Zhou/MRM}}
\endgroup

\begin{abstract}
Unsupervised domain adaptation (UDA) for semantic segmentation seeks to transfer models from a labeled source domain to an unlabeled target domain. While auxiliary self-supervised tasks such as contrastive learning have enhanced feature discriminability, masked modeling remains underexplored due to architectural constraints and misaligned objectives. We propose Masked Representation Modeling (MRM), an auxiliary task that performs representation masking and reconstruction directly in the latent space. Unlike prior masked modeling methods that reconstruct low-level signals (e.g., pixels or visual tokens), MRM targets high-level semantic features, aligning its objective with segmentation and integrating seamlessly into standard architectures like DeepLab and DAFormer. To support efficient reconstruction, we design a lightweight auxiliary module, Rebuilder, which is jointly trained with the segmentation network but removed during inference, introducing zero test-time overhead. Extensive experiments demonstrate that MRM consistently improves segmentation performance across diverse architectures and UDA benchmarks. When integrated with four representative baselines, MRM achieves an average gain of +2.3 mIoU on GTA $\rightarrow$ Cityscapes and +2.8 mIoU on Cityscapes $\rightarrow$ Synthia, establishing it as a simple, effective, and generalizable strategy for unsupervised domain-adaptive semantic segmentation. 
\end{abstract}    
\section{Introduction}
\label{sec:intro}

Deep learning has achieved significant success in computer vision, including tasks such as semantic segmentation ~\cite{minaee2021image,sung2024contextrast,ni2024context}. However, these models often struggle when faced with domain shift—a mismatch between training and testing data distributions—which can lead to notable performance degradation \cite{wang2018deep}. Addressing this issue by collecting and annotating new target domain data is resource-intensive and time-consuming, particularly for dense pixel-level annotations in semantic segmentation \cite{cordts2016cityscapes,sakaridis2021acdc}. Unsupervised domain adaptation (UDA) \cite{zhou2024unsupervised} offers a practical alternative, enabling models to utilize labeled source domain data and unlabeled target domain data to improve performance without additional annotation efforts.

\begin{figure}[!t]
	\centering
	\includegraphics[width=3.3in]{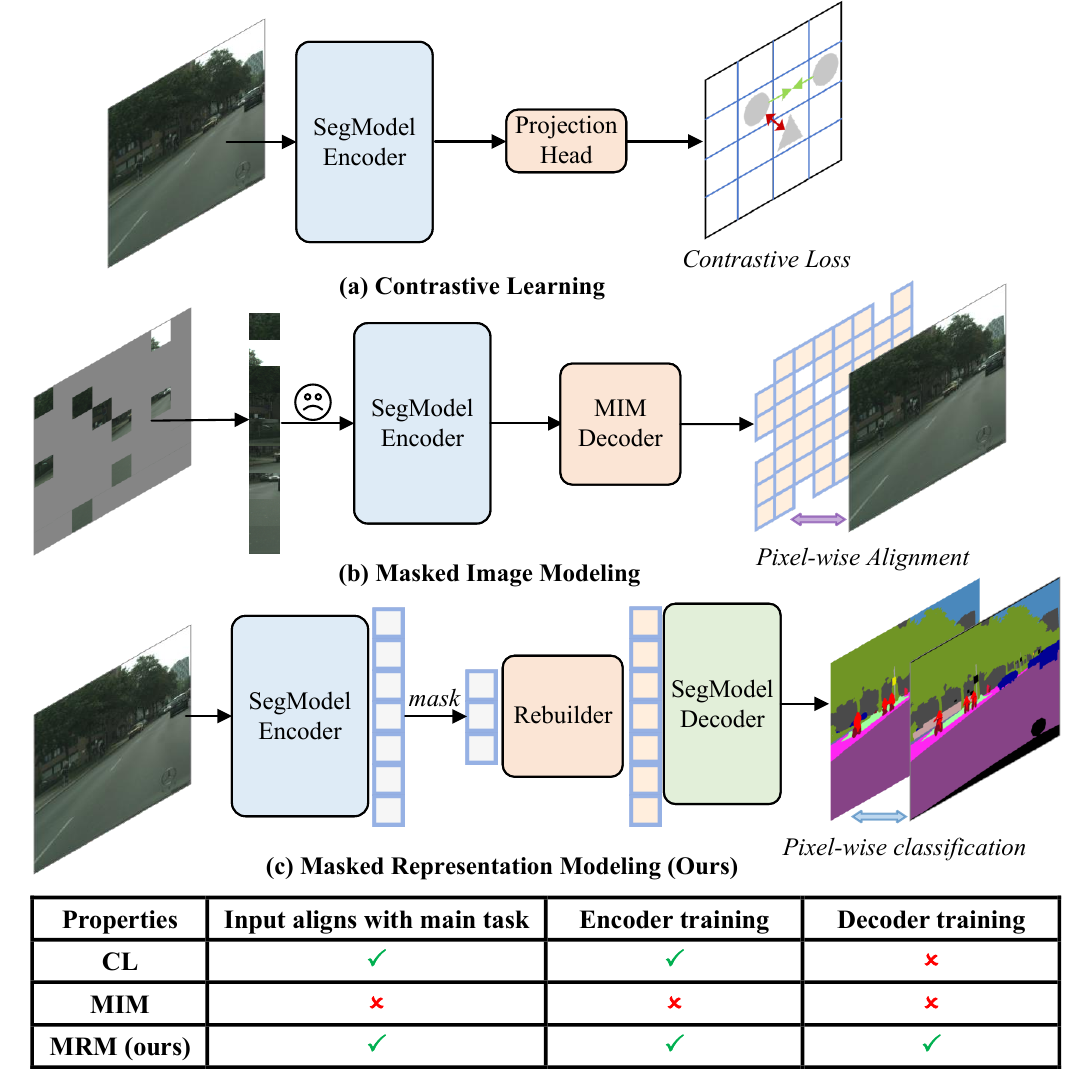}
	\caption{Comparison of three auxiliary tasks for UDA segmentation. (a) Contrastive Learning (CL) uses contrastive loss for feature alignment but does not train the decoder, limiting end-to-end optimization. (b) Masked Image Modeling (MIM) reconstructs masked components but disrupts the segmentation pipeline, reducing compatibility with certain architectures. (c) Masked Representation Modeling (MRM) performs masking and reconstruction in latent space, aligns with the segmentation task, remains compatible with diverse architectures, and improves performance without inference overhead, as the Rebuilder is used only during training.}
	\label{fig1}
\end{figure}

Auxiliary tasks provide a promising direction for enhancing UDA-based segmentation models ~\cite{xie2023sepico,chen2023pipa,tsai2018learning, hoyer2023improving}. These tasks, typically designed to require no manual annotations and to integrate seamlessly with existing architectures, enrich representations without disrupting the primary network. Among them, contrastive learning \cite{chen2020simple, he2020momentum, oquab2023dinov2, simeoni2025dinov3} has shown strong potential by improving feature discrimination: pulling similar samples closer while pushing dissimilar ones apart (Figure~\ref{fig1}(a)). This makes it particularly useful for UDA segmentation, where distinguishing challenging categories is critical \cite{xie2023sepico,chen2023pipa,lee2022bi,li2023contrast, chen2024dynamic}. However, another self-supervised technique—masked image modeling (MIM)—has seen limited exploration in the context of UDA segmentation. MIM methods, such as masked autoencoders (MAE) \cite{he2022masked}, train models to reconstruct occluded regions, fostering a better understanding of global context and scene structure. While this approach seems suitable for UDA segmentation by encouraging models to capture broader scene context, its adoption is limited. We speculate that this mainly stems from two main challenges:

\textbf{(i)} \textit{Input structure constraints:} MIM modifies the input structure by masking image patches (Figure~\ref{fig1}(b)), complicating its application to segmentation networks like DeepLab \cite{chen2017deeplab} or DAFormer \cite{hoyer2022daformer}. 

\textbf{(ii)} \textit{Optimization conflicts:} MIM methods focus on element-wise reconstructing occluded patches, which may introduce conflicts with the optimization objectives of domain adaptive segmentation tasks. 

To address these challenges, we propose Masked Representation Modeling (MRM), a simple yet effective auxiliary task for unsupervised domain-adaptive semantic segmentation (Figure~\ref{fig1}(c)). Unlike image-level masking methods, MRM performs representation masking and reconstruction in latent space, avoiding disruptions to the input and making it compatible with a wide range of architectures, including CNNs \cite{chen2017deeplab} and Transformers \cite{hoyer2022daformer}. More importantly, MRM aligns its optimization objective with the primary segmentation task by using the segmentation decoder to perform pixel-wise classification on reconstructed representation, rather than enforcing alignment in pixel space \cite{he2022masked, woo2023convnext}, thereby reducing conflicts between auxiliary and main tasks and uniquely enhancing the decoder—a benefit not typically offered by contrastive learning \cite{chen2020simple, he2020momentum}. By encouraging the network to predict missing latent representations, MRM implicitly improves feature robustness and cross-domain generalization, which is particularly critical in scenarios with large domain shifts.

To facilitate representation reconstruction, we introduce a lightweight, asymmetric Rebuilder module inspired by masked image modeling \cite{he2022masked, assran2023self, wang2023masked, woo2023convnext}. The Rebuilder is jointly trained with the segmentation network and removed after training, introducing zero additional inference overhead. Extensive experiments demonstrate that MRM consistently improves segmentation performance across diverse architectures and benchmarks. For instance, when combined with four representative baselines, it achieves average gains of +2.3 mIoU on GTA→Cityscapes and +2.8 mIoU on Cityscapes→Synthia, highlighting MRM as a generalizable, plug-and-play strategy for enhancing UDA segmentation. These results suggest that representation-level masked modeling can serve as a versatile auxiliary objective, complementing existing adaptation techniques without requiring architectural modifications or complex training procedures.

\section{Related Work}
\label{sec:related_work}

\textbf{Semantic segmentation}, a cornerstone of computer vision, has seen significant progress with deep learning. Fully Convolutional Networks (FCNs) \cite{long2015fully} enabled pixel-level understanding, but challenges like small object delineation and complex scenes persist. The encoder-decoder architecture improved segmentation with innovations like skip-connections \cite{ronneberger2015u} for feature fusion and dilated convolutions \cite{chen2018encoder} to expand receptive fields. While CNN-based methods dominated early research, Transformers \cite{zheng2021rethinking,xie2021segformer,cheng2022masked,kerssies2025your, zhou2025nnwnet} have advanced global context modeling. Despite their performance, Transformers' self-attention mechanism is computationally intensive for high-resolution images \cite{vaswani2017attention}. Xie \textit{et al.} \cite{xie2021segformer} mitigated this by downsampling key and value components in self-attention.

\textbf{Unsupervised domain adaptive segmentation} addresses the challenge of adapting segmentation models to new domains without target annotations. Common approaches include adversarial training \cite{tsai2018learning, hoffman2018cycada}, self-training \cite{tranheden2021dacs, zou2018unsupervised, vuong2025preserving}, and efficient architecture design \cite{hoyer2022daformer, hoyer2024domain}. Beyond these, auxiliary tasks have been introduced to enhance feature representation: contrastive learning improves backbone features \cite{xie2023sepico, chen2023pipa, assefa2025dycon}, while depth estimation aids scene understanding \cite{hoyer2023improving, yang2024micdrop}. Though effective, these methods typically focus on the encoder and may conflict with the main task. Our method, Masked Representation Modeling (MRM), overcomes these limitations by training the entire model, aligning encoder and decoder under the same objective to reduce potential conflicts.

\textbf{Masked image modeling (MIM)} learns visual representations by reconstructing masked image regions \cite{zhang2025linguistics, hermosilla2025masked, lin2025prototypes, he2022masked, woo2023convnext}. Originating from image inpainting, Context Encoders \cite{pathak2016context} used convolutional networks to predict missing regions. Building on this idea, ViT-based approaches \cite{Dosovitskiy2021vit, bao2021beit, he2022masked} extended masking to self-supervised learning, from patch prediction to discrete token modeling, while ConvNeXtV2 \cite{woo2023convnext} demonstrated its effectiveness for convolutional architectures. Despite the success of MIM in representation learning, masking input patches disrupts the data-processing pipeline, making methods like MAE \cite{he2022masked} difficult to integrate into segmentation frameworks such as DeepLab \cite{chen2017deeplab} and DAFormer \cite{hoyer2022daformer}. To address this, we propose MRM, which performs masking and reconstruction in the latent space, preserving the input paradigm and supporting diverse segmentation architectures.

\section{Method}
\label{sec:methods}

\subsection{Preliminary}
\textbf{Unsupervised domain adaptive segmentation} aims to train a neural network using labeled source domain data  $D_{s}=\left\{\left(x_{k}^{s}, y_{k}^{s}\right)\right\}_{k=1}^{n_{s}}$ to perform well on a target domain $D_{t}=\left\{\left(x_{k}^{t}\right)\right\}_{k=1}^{n_{t}}$, without access to target labels. In semantic segmentation model (SegModel), typically consists of an Encoder $E(\cdot)$ and a Decoder $D(\cdot)$. Simply training the network using pixel-wise cross-entropy on the source domain can be formulated as:
\begin{eqnarray}
	\mathcal{L}_{sup}=-\sum_{i=1}^{H} \sum_{j=1}^{W} \sum_{c=1}^{C} y^{s}_{i j c} \log D(E(x^s))_{i j c}
\end{eqnarray}
where $H$ and $W$ represent the image's height and width respectively, while $C$ signifies the number of categories in the UDA task.

However, a model trained solely on the source domain often experiences a performance decline when applied to a different domain. To address this issue, UDA methods leverage unlabeled target domain images to adapt the network. To achieve this, an additional unsupervised loss $\mathcal{L}_{uda}$ is introduced into the optimization process.
\begin{eqnarray}
	\mathcal{L}_{overall} = \mathcal{L}_{sup} + \mathcal{L}_{uda}
\end{eqnarray}

\textbf{Masked image modeling} is a self-supervised learning paradigm designed to train neural networks by reconstructing masked portions of an input image. The input image $x$ is reshaped into non-overlapping patches $p=\left\{p_{i}\right\}_{i=1}^{N_l}$. MIM constructs a random mask $M^{mim} \in\{0,1\}^{N_l}$ to indicate the masked patches, where $M^{mim}_{i}=1$ corresponds to the patches that are masked. Only the visible patches $p^{v}=\left\{p_{i} \mid M^{mim}_{i}=0\right\}_{i=1}^{N_v}$ are fed into the encoder and are mapped to potential features, after which a lightweight decoder $D_{mim}$ is introduced to reconstruct the pixel values of the invisible patches. 
\begin{eqnarray}
	\mathcal{L}_{mim}=\sum_{i=1}^{N_{l}} M^{mim}_{i} \cdot \left ( D_{mim}(E(p^{v}_{i})) - p_{i} \right )^{2}
\end{eqnarray}

Incorporating MIM as an auxiliary task poses challenges since typical UDA segmentation architectures \cite{chen2017deeplab, hoyer2022daformer} are designed to process complete images rather than visible patches.

\subsection{Masked Representation Modeling}

Masked Representation Modeling (MRM) is a simple approach designed to address the limitations of mainstream MIM methods when used as an auxiliary task. Unlike MIM, MRM eliminates the need for the SegModel encoder $E(\cdot)$ to process partially visible signals. Instead, it reconstructs the latent representation from the partially visible features provided to the encoder. The SegModel decoder $D(\cdot)$ then processes the reconstructed representation, and the model is trained through a pixel-wise classification task.

\begin{eqnarray}
	\mathcal{L}_{mrm}=-\sum_{i=1}^{H} \sum_{j=1}^{W} \sum_{c=1}^{C} \tilde{y}_{i j c} \log D(R(E(x^t)))_{i j c}
\end{eqnarray}

Here, $\tilde{y}$ is the pseudo label \cite{lee2013pseudo} of target image, and $R(\cdot)$ represents the Rebuilder (\textit{refer to next section for detailed information}), whose role is analogous to the lightweight decoder in MIM \cite{he2022masked,woo2023convnext}. Both are responsible for reconstructing the original signal from partially visible signals. When the MRM task is used as an auxiliary task for unsupervised domain adaptive segmentation, the overall optimization objective can be expressed as:

\begin{eqnarray}
	\mathcal{L}_{overall} = \mathcal{L}_{sup} + \mathcal{L}_{uda} + \lambda\mathcal{L}_{mRm}
\end{eqnarray}
where $\lambda$ is the trade-off in masked representation modeling. MRM can be easily integrated with existing UDA methods.

\subsection{Rebuilder}

The Rebuilder $R(\cdot)$ primarily serves to randomly mask out representation regions from SegModel encoder and then reconstruct them. Its design is largely consistent with the decoder \cite{he2022masked, wang2023masked, assran2023self, bardes2024revisiting} used in MIM. After the semantic segmentation model has been fully trained, the Rebuilder is removed, ensuring it does not impact the inference process of the original model. The pipeline of the Rebuilder is shown in Figure~\ref{fig2}, and the following sections will discuss the specific design details. See \textbf{Supplementary Material} for further analysis.

\begin{figure*}[!t]
	\centering
	\includegraphics[width=6.8in]{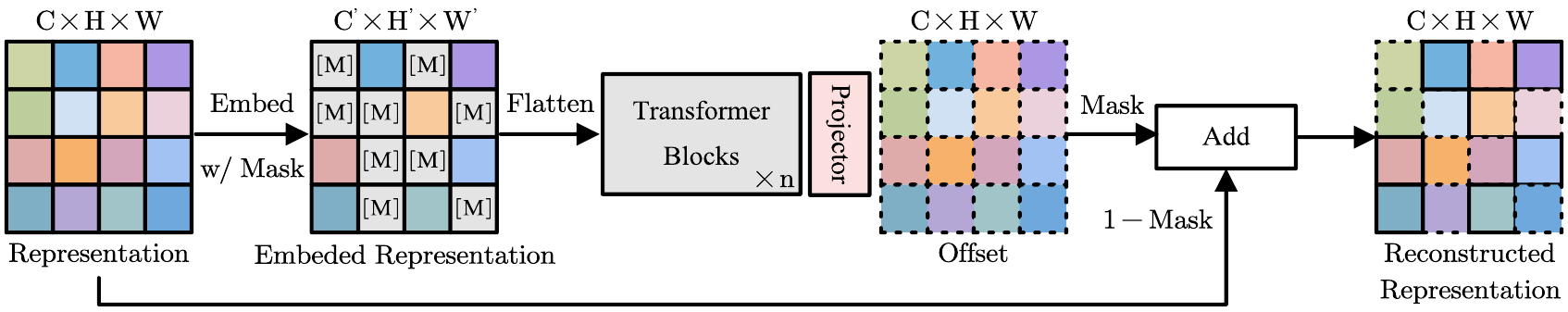}
	\caption{The pipeline of Rebuilder. The Rebuilder is designed to randomly mask out representation from the encoder and reconstruct the masked component. It first scales the encoder representation along both spatial and channel dimensions, and then applies random masking to remove a subset of these representation. Subsequently, the masked representation are passed through several Transformer blocks and a projector to generate reconstructed representation, which are input to the decoder for model training. [M] is a learnable token.} 
	\label{fig2}
\end{figure*}

\textbf{Representation embedding.} Since different encoders \cite{chen2017deeplab, hoyer2022daformer} produce representation at varying scales, it is essential to ensure that subsequent transformer blocks can efficiently process the features from different encoders in a consistent manner. Thus, the input feature $f^t=E(x^t), f^t\in {{\mathbb{R}}^{C \times H \times W}}$ is scaled to $C^\prime \times H^\prime \times W^\prime$ to obtain embedding representation $\tilde{f}^t$. For the channel dimension, a linear mapping layer is applied for rescaling. After processing the channel dimension, bilinear interpolation is used to resize the input representation to the target spatial dimensions when the input and target spatial dimensions are not aligned. 

\textbf{Masking.} Based on the shape of the features obtained after representation embedding, a binary mask $M \in {{\mathbb{R}}^{H^\prime \times W^\prime}}$ is generated through uniform random sampling \cite{he2022masked}. We apply the generated mask to perform masked out operation on the embedding representation, and fill the corresponding positions of the removed representation with a mask token. Each mask token \cite{kenton2019bert} is a shared, learnable vector that represents the presence of a missing patch to be predicted. The processed representation are reshaped to $(H^\prime W^\prime) \times C^\prime$ and fed into the subsequent parts of the architecture.

\begin{figure}[!t]
	\centering
	\includegraphics[width=2.9in]{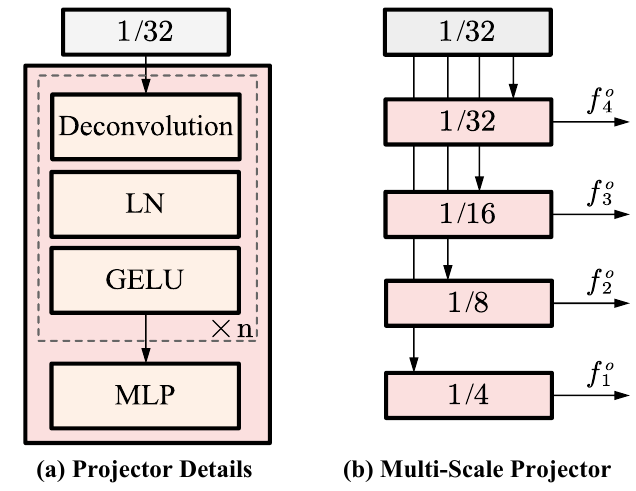}
	\caption{An overview of the projector. The representation from the Transformer are reshaped and fed into the projector, which uses several transposed convolutions to generate features at different scales. (a) The projector details and (b) the multi-scale projector.}
	\label{fig3}
\end{figure}

\textbf{Architecture.} Since the mask token lacks positional information, following works \cite{he2022masked, wang2023masked, assran2023self, bardes2024revisiting} the reshaped representation are first added with absolute positional embeddings \cite{vaswani2017attention}. A series of Transformer blocks \cite{Dosovitskiy2021vit} are then applied to process these representation. Experiment demonstrate that MRM achieves performance gains with only a minimal number of blocks (e.g., $n=1$ or $n=2$), thus avoiding significant computational overhead, details in Table~\ref{tab2}(b). In addition, a projector is used to map the Transformer’s output to the original representation dimensions. The projector consists of transposed convolution layers that rescale both spatial and channel dimensions to align with the original representation, as shown in Figure~\ref{fig3}(a). 

After obtaining the offset $f^o$ from the Projector's output, the mask is resized to the original feature map size to obtain $M^s \in {{\mathbb{R}}^{H \times W}}$. Based on $M^s$, the offset are fused with the original representation $f^t$ to obtain the final reconstructed representation $f^r$, where broadcasting is applied due to the channel dimension mismatch between the mask and feature tensors.

\begin{eqnarray}
	f^{r} = M^{s} \odot f^{o} +(1-M^{s}) \odot f^{t}
\end{eqnarray}
where $\odot$ is Hadamard product.

\textbf{Implemented with multi-scale model.} For multi-scale models (e.g., DAFormer \cite{hoyer2022daformer}), we do not instantiate a separate Rebuilder at each stage, as such a naive design would introduce substantial computational and memory overhead due to the Transformer components. Instead, we leverage only the representation from the final encoder stage for embedding and Transformer processing. As illustrated in Figure~\ref{fig3}(b), we introduce an individual upsampling operation for each target scale to project the final-stage representation into multi-scale features. In this way, the reconstructed features at different resolutions are generated directly from the same high-level representation, rather than by applying Transformer-based rebuilding at every stage. This design preserves the multi-scale property of hierarchical architectures while avoiding redundant computation. The feasibility of this design is supported by ViTDet \cite{li2022exploring}, which shows that multi-scale features can be derived from final-stage representations through simple upsampling operations. Consequently, for architectures such as DAFormer \cite{hoyer2022daformer}, our MRM introduces no additional Transformer overhead while maintaining effective reconstruction across scales. Formally, given a hierarchical representation $f^{t}_{i}$, where $i \in {1,2,3,4}$, the reconstructed representation is computed as

\begin{eqnarray}
	f^{r}_{i} = M^{s}_{i} \odot f^{o}_{i} +(1-M^{s}_{i}) \odot f^{t}_{i}.
\end{eqnarray}
\section{Experiments}
\label{sec:experiments}

\definecolor{lightgray}{gray}{0.93}

\begin{table*}[!t]
	\centering
	\small
	\setlength{\tabcolsep}{0.65pt}
	\begin{tabular}{c|ccccccccccccccccccc|c}
		\toprule
		Method &Road &S.walk &Build. &Wall &Fence &Pole &Tr.Light &Sign &Veget. &Terrain &Sky &Person &Rider &Car &Truck &Bus &Train &M.bike &Bike &mIOU\\
		\midrule
		\multicolumn{21}{c}{GTA $\rightarrow$ Cityscape}\\
		\midrule
		PiPa \cite{chen2023pipa} & 96.1 & 72.0 & 90.3 & 56.6 & 52.0 & 55.1 & 61.8 & 63.7 & 90.8 & 52.6 & 93.6 & 74.3 & 43.6 & 93.5 & 78.4 & 84.2 & 77.3 & 59.9 & 66.7 & 71.7 \\
		GANDA \cite{liao2023geometry} & 96.5 & 74.8 & 91.4 & 61.7 & 57.3 & 59.2 & 65.4 & 68.8 & 91.5 & 49.9 & \textbf{94.7} & 79.6 & 54.8 & 94.1 & 81.3 & 86.8 & 74.6 & 64.8 & 68.2 & 74.5 \\
		QuadMix \cite{zhang2025unified} & 97.5 & \textbf{80.9} & 91.6 & 62.3 & \textbf{57.6} & 58.2 & 64.5 & 71.2 & 91.7 & 52.3 & 94.3 & 80.0 & 55.9 & \textbf{94.6} & 86.3 & 90.5 & 82.3 & 65.1 & 68.1 & 76.1\\
		\midrule
		DACS \cite{tranheden2021dacs}& 89.9& 39.7& 87.9& 30.7& 39.5& 38.5& 46.4& 52.8& 88.0& 44.0& 88.8& 67.2& 35.8& 84.5& 45.7& 50.2& 0.0& 27.3& 34.0& 52.1\\
		\rowcolor{lightgray}
		w/ MRM & 94.7 & 68.3 & 87.6 & 38.1 & 27.4 & 42.0 & 51.9 & 59.2 & 86.9 & 45.0 & 87.7 & 66.3 & 32.3 & 89.6 & 53.9 & 56.6 & 0.0 & 33.2 & 42.1 & 55.9\\
		DAFormer \cite{hoyer2022daformer}&  95.7 &  70.2 &  89.4 &  53.5 &  48.1 &  49.6 &  55.8 &  59.4 &  89.9 &  47.9 &  92.5 & 72.2 & 44.7 & 92.3 & 74.5 &  78.2 &  65.1 &  55.9 &  61.8 &  68.3\\
		\rowcolor{lightgray}
		w/ MRM & 96.8 & 76.3 & 89.5 &  55.7 & 50.3 & 50.9 & 58.2 & 62.2 & 90.0 & 50.1 & 91.1 & 73.8 & 48.2 & 92.3 & 77.3 & 80.4 & 71.6 & 56.3 & 64.0  & 70.3 \\
		HRDA \cite{hoyer2024domain}& 96.4& 74.4& 91.0& 61.6& 51.5& 57.1& 63.9& 69.3& 91.3& 48.4& 94.2& 79.0& 52.9& 93.9& 84.1& 85.7& 75.9& 63.9& 67.5& 73.8\\
		\rowcolor{lightgray}
		w/ MRM & 97.0 & 78.3 & 90.9 & 59.5 & 52.8 & 60.4 & 66.6 & 72.4 & \textbf{91.9} & 51.8 & 94.3 & 79.3 & 55.5 & 94.4 & 84.4 & 87.9 & 82.7 & 65.4 & 68.2 & 75.4\\
		MIC \cite{hoyer2023mic}& 97.4& 80.1& 91.7& 61.2& 56.9& 59.7& 66.0& 71.3& 91.7& 51.4& 94.3& 79.8& 56.1& \textbf{94.6}& 85.4& 90.3& 80.4& 64.5& 68.5& 75.9\\
		\rowcolor{lightgray}
		w/ MRM & \textbf{98.3} & 80.4 & \textbf{92.6} & \textbf{62.7} & 57.0 & \textbf{62.3} & \textbf{69.1} & \textbf{74.3} & 91.8 & \textbf{53.5} & \textbf{94.7} & \textbf{81.1} & \textbf{56.6} & 94.1 & \textbf{87.2} & \textbf{91.6} & \textbf{85.4} & \textbf{66.3} & \textbf{71.3} & \textbf{77.5} \\
		\midrule
		\multicolumn{21}{c}{Synthia $\rightarrow$ Cityscape}\\
		\midrule
		PiPa \cite{chen2023pipa} & 87.9 & 48.9 & 88.7 & 45.1 & 4.5 & 53.1 & 59.1 & 58.8 & 87.8 & – & 92.2 & 75.7 & 49.6 & 88.8 & – & 53.5 & – & 58.0 & 62.8 & 63.4 \\
		GANDA \cite{liao2023geometry} &  89.1 & 50.6 & \textbf{89.7} & \textbf{51.4} & 6.7 & 59.4 & \textbf{66.8} & 57.7 & 86.7 & – & 93.8 & 80.6 & 56.9 & \textbf{90.7} & – & 64.8 & – & 62.6 & 65.0 & 67.0 \\
		QuadMix \cite{zhang2025unified} & 88.1 & 51.2 & 88.9 & 46.7 & 7.9 & 58.6 & 64.7 & \textbf{63.7} & 88.1 & – & 93.9 & 81.3 & 56.6 & 90.3 & – & \textbf{66.9} & – & 66.8 & \textbf{66.0} & 67.5\\
		\midrule
		DACS \cite{tranheden2021dacs} &80.6 &25.1 &81.9 &21.5& 2.9 &37.2 &22.7 &24.0 &83.7 &– & 90.8 &67.6 &38.3 &82.9 &– &38.9 &– &28.5 &47.6 &48.3\\
		\rowcolor{lightgray}
		w/ MRM & 83.5 & 44.2 & 84.9 & 21.5 & 3.4 & 41.6 & 47.2 & 52.3 & 83.7 & – & 86.5 & 68.6 & 42.1 & 84.1 & – & 51.2 & – & 40.4 & 58.2 & 55.8 \\
		DAFormer \cite{hoyer2022daformer}& 84.5& 40.7& 88.4& 41.5& 6.5& 50.0& 55.0& 54.6& 86.0& –& 89.8& 73.2& 48.2& 87.2& –& 53.2& –& 53.9& 61.7& 60.9\\
		\rowcolor{lightgray}
		w/ MRM & 88.4 & 51.4 & 88.8 & 41.2 &  \textbf{8.4} & 50.2 & 55.9 & 52.9 &	85.8 & – & 88.5 & 73.0 & 47.6 & 87.2 & – & 63.1 & – & 57.8 & 61.4 & 62.6 \\
		HRDA \cite{hoyer2024domain} & 85.2& 47.7& 88.8& 49.5& 4.8& 57.2& 65.7& 60.9& 85.3& –& 92.9& 79.4& 52.8& 89.0& –& 64.7& –& 63.9& 64.9& 65.8\\
		\rowcolor{lightgray}
		w/ MRM & \textbf{90.6} & \textbf{55.5} & 88.3 & 49.8 &  7.0 & 57.8 & 65.6 & 56.6 & 87.9 & – & 93.9 & 79.6 & 53.2 & 89.1 & – & 65.1 & – & \textbf{67.8} & \textbf{66.0} & 67.1 \\
		MIC \cite{hoyer2023mic}& 86.6& 50.5& 89.3 & 47.9& 7.8& 59.4& 66.7 & 63.4& 87.1 & – & 94.6 & 81.0& 58.9& 90.1& –& 61.9& –& 67.1& 64.3& 67.3\\
		\rowcolor{lightgray}
		w/ MRM & 87.5 & 53.2 & \textbf{89.7} & 48.7 &  7.8 & \textbf{61.0} & 66.7 & 63.1 & \textbf{88.6} &  – & \textbf{94.7} & \textbf{81.8} & \textbf{59.7} & 90.1 & – & 64.3 &  – & \textbf{67.8} & 65.3  & \textbf{68.1} \\
		\bottomrule
	\end{tabular}%
	\caption{Comparison with state-of-the-art methods for UDA. The best result in each metric column is marked bold. The results for MRM are averaged over 3 random seeds}
	\label{tab1}%
\end{table*}%

In this section, we focus on experimenting with two popular benchmarks: GTA $\rightarrow$ Cityscape and Synthia $\rightarrow$ Cityscape. First, we introduce the datasets along with the implementation details. Then, we compare our approach with state-of-the-art (SoTA) methods. Next, we explore the performance impact under various parameter settings through ablation studies. Finally, we present qualitative analysis to further illustrate the effectiveness of our method.

\subsection{Experiment Setups}
\textbf{Dataset.} The GTA \cite{richter2016playing} dataset comprises 24,966 synthetic images featuring pixel-level semantic annotations. These images are generated within the open-world environment of ``Grand Theft Auto V'', all captured from the perspective of a vehicle navigating the streets of American-style virtual cities. This dataset encompasses 19 semantic classes that align with those found in the Cityscapes dataset.

SYNTHIA \cite{ros2016synthia} constitutes a synthetic urban scene dataset. We opt for its subset known as ``SYNTHIA-RAND-CITYSCAPES'', which shares 16 common semantic annotations with Cityscapes. Specifically, we utilize a total of 9,400 images, each with a resolution of 1280×760, sourced from the SYNTHIA dataset.

Cityscapes \cite{cordts2016cityscapes} is a dataset featuring real urban scenes captured across 50 cities in Germany and neighboring regions. The dataset includes meticulously annotated images, comprising 2,975 training images, 500 validation images, and 1,525 test images, all at a resolution of 2048×1024 pixels. Each pixel within these images is classified into one of 19 distinct categories.

\textbf{Baseline methods.} We select popular UDA segmentation methods such as DACS \cite{tranheden2021dacs}, DAFormer \cite{hoyer2022daformer}, HRDA \cite{hoyer2024domain}, and MIC \cite{hoyer2023mic} as baselines to validate the performance improvement of MRM on these baselines. We also compare our method with popular approaches such as QuadMix \cite{zhang2025unified}, PiPa \cite{chen2023pipa}, and GANDA \cite{liao2023geometry}.

\textbf{Implementation details.} To ensure fairness and reproducibility, the hyperparameters related to the Baselines are kept consistent with those in the original paper. For network architecture, we utilize the DeepLab-V2 \cite{chen2017deeplab} with ResNet101 \cite{he2016deep} and DAFormer \cite{hoyer2022daformer} with MiT-B5 \cite{xie2021segformer} as the architecture. 

For Masked Representation Modeling, we chose a trade-off $\lambda$ value of 1.0. Regarding the design of the Rebuilder, the number of transformer blocks is set to 2, and the embedding dimension is set to 512. In the representation embedding section, we choose to scale the spatial and channel dimensions of the feature map, specifically setting $H^\prime =W^\prime =16$ and $C^\prime =512$. We set the masking ratio to 40\%. The learning rate of $2 \times 10^{-4}$ is employed with the Rebuilder. For the HRDA head \cite{hoyer2024domain}, the MRM is trained using both the fused multi-scale predictions and the detail crop predictions, following the methodology described in the original paper. The results are averaged over 3 random seeds. All the experiments are conducted with 1x NVIDIA GTX 3090 with 24G RAM and the PyTorch framework is implemented to perform our experiments.

\subsection{Comparisons with State-of-the-Arts UDA Methods}

We evaluate our method against SoTA UDA approaches on two widely used benchmarks: GTA5 $\rightarrow$ Cityscapes and SYNTHIA $\rightarrow$ Cityscapes, using a variety of base learners. As shown in Table~\ref{tab1}, integrating the proposed MRM module leads to consistent performance improvements.

On the GTA5 $\rightarrow$ Cityscapes benchmark, MRM yields notable gains: +3.8 mIoU for DACS \cite{tranheden2021dacs}, +2.0 mIoU for DAFormer \cite{hoyer2022daformer}, +1.6 mIoU for HRDA \cite{hoyer2024domain}, and +1.6 mIoU for MIC \cite{hoyer2023mic}. The improvement is particularly pronounced in fine-grained categories such as traffic sign, rider, and motorbike, suggesting that MRM may enhance the decoder’s capacity for high-level semantic discrimination. In particular, the MIC \cite{hoyer2023mic} + MRM combination achieves an mIoU of 77.5, which surpasses previous state-of-the-art results by a margin of +1.4.

On the more challenging SYNTHIA $\rightarrow$ Cityscapes benchmark, MRM consistently improves performance: +7.5 mIoU for DACS \cite{tranheden2021dacs}, +1.7 for DAFormer \cite{hoyer2022daformer}, +1.3 for HRDA \cite{hoyer2024domain}, and +0.8 for MIC \cite{hoyer2023mic}. These results suggest that MRM generalizes well across different adaptation strategies and datasets.

Overall, the findings indicate that MRM serves as a model-agnostic, plug-and-play auxiliary task that can complement existing UDA pipelines by more effectively aligning feature learning with semantic segmentation objectives.

\subsection{Ablation Study}

In the ablation study, we adopt DACS \cite{tranheden2021dacs} as the baseline and conduct experiments on the GTA $\rightarrow$ Cityscapes benchmark using both DeepLab-V2 \cite{chen2017deeplab} with ResNet101 \cite{he2016deep} and DAFormer \cite{hoyer2022daformer} with MiT-B5  \cite{xie2021segformer} architectures to ensure consistent and generalizable analysis across different designs.

\begin{figure}[!t]
	\centering
	\includegraphics[width=2.9in]{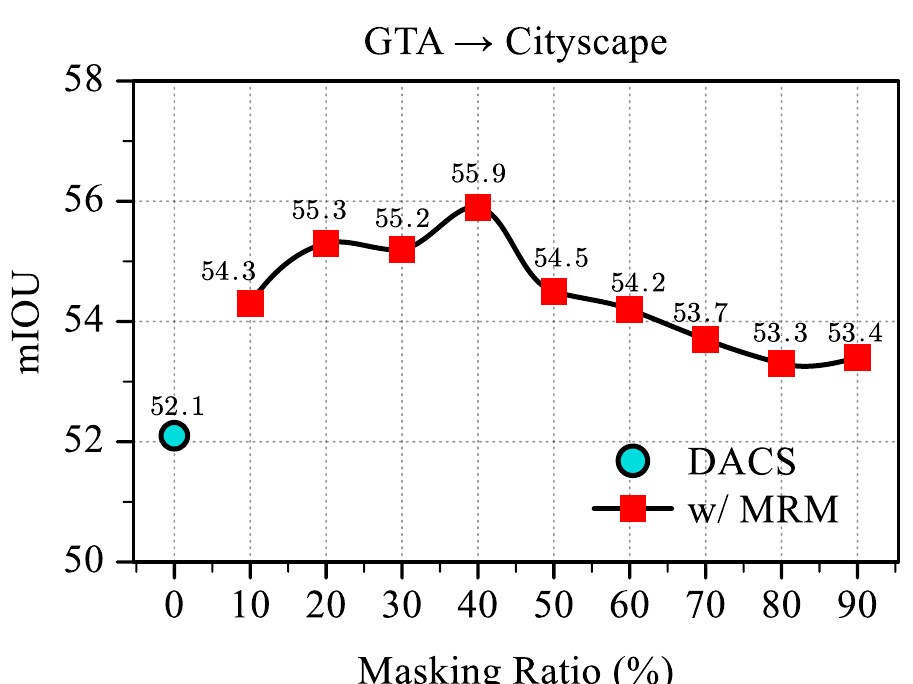}
	\caption{Masking ratio. The optimal performance enhancement is achieved when the masking ratio is adjusted to 40\%.}
	\label{fig4}
\end{figure}

\textbf{Masking ratio.} Figure~\ref{fig4} shows that MRM performs best at a masking ratio of 40\%, reaching 55.9 mIoU (+3.8 over baseline). Unlike MAE \cite{he2022masked} and ConvNeXtV2 \cite{woo2023convnext}, which favor high masking ratios (60–75\%), MRM benefits from lower ratios. Excessive masking reduces the diversity of reconstructed representation and harms semantic consistency, especially since MRM's visible tokens are processed by fewer Transformer blocks than in full MIM setups.

\begin{table*}[!t]
	\centering
	\small
	\subfloat[
	Embedding dimensions of Transformer blocks in the Rebuilder.
	]{
		\centering
		\begin{minipage}{0.29\linewidth}{\begin{center}
					\begin{tabular}{ccccc}
						\toprule
						Dim & 128 & 256 & 512 & 768\\
						\midrule
						mIOU & 54.7 & 55.1 & \textbf{55.9} & 53.6 \\
						\bottomrule
						\multicolumn{4}{c}{~}\\
						\multicolumn{4}{c}{~}\\
					\end{tabular}
		\end{center}}\end{minipage}
	}
	\hfill
	\subfloat[
	Number of Transformer blocks used in the Rebuilder.
	]{
		\centering
		\begin{minipage}{0.29\linewidth}{\begin{center}
					\begin{tabular}{ccccc}
						\toprule
						Blocks & 1 & 2 & 4 & 8  \\
						\midrule
						mIOU & 55.4 & \textbf{55.9} & 54.4 & 52.9 \\
						\bottomrule
						\multicolumn{4}{c}{~}\\
						\multicolumn{4}{c}{~}\\
					\end{tabular}
		\end{center}}\end{minipage}
	}
	\hfill
	\subfloat[
	Effect of masking and rebuilding on performance.
	]{
		\centering
		\begin{minipage}{0.29\linewidth}{\begin{center}
					\renewcommand{\arraystretch}{0.88}
					\begin{tabular}{c|c}
						\toprule
						Case & mIOU \\
						\midrule
						Baseline & 52.1 \\
						\midrule
						Masking & 51.9 \\
						Masking + Rebuilding & \textbf{55.9} \\
						\bottomrule
					\end{tabular}
		\end{center}}\end{minipage}
	}
	\vspace{2.0em}
	
	\subfloat[
	Spatial dimensions of the rescaled representation. OOM marks out of memory.
	]{
		\centering
		\begin{minipage}{0.29\linewidth}{\begin{center}
					\setlength{\tabcolsep}{4.0pt}
					\begin{tabular}{ccccc}
						\toprule
						$H^\prime = W^\prime$ & 8 & 16 & 32 & 64\\
						\midrule
						mIOU & 55.6  &\textbf{55.9} & 54.1 & OOM \\
						\bottomrule
						\multicolumn{4}{c}{~}\\
						\multicolumn{4}{c}{~}\\
						\multicolumn{4}{c}{~}\\
						\multicolumn{4}{c}{~}\\
					\end{tabular}
		\end{center}}\end{minipage}
	}
	\hfill
	\subfloat[
	Training objective. Consistent objectives boost performance.
	]{
		\centering
		\begin{minipage}{0.29\linewidth}{\begin{center}
					\renewcommand{\arraystretch}{0.92}
					\begin{tabular}{c|c}
						\toprule
						Case & mIOU \\
						\midrule
						Baseline & 52.1 \\
						\midrule
						pixel rec. (w/ norm) \cite{he2022masked} & 51.8 \\
						feature rec. (w/ teacher) \cite{bardes2024revisiting} & 53.5 \\
						feature rec. (w/o teacher) \cite{bardes2024revisiting} & 53.7 \\
						pixel cls. & \textbf{55.9} \\
						\bottomrule
					\end{tabular}
		\end{center}}\end{minipage}
	}
	\hfill
	\subfloat[
	Influence of the loss weight $\lambda$ on performance.
	]{
		\begin{minipage}{0.29\linewidth}{\begin{center}
					\centering
					\setlength{\tabcolsep}{3.2pt}
					\begin{tabular}{cccccc}
						\toprule
						$\lambda$ & 0.1 & 0.5 & 1.0 & 2.0 & 10.0 \\
						\midrule
						mIOU & 54.7 &55.8  & \textbf{55.9} & 55.4  & 52.1 \\
						\bottomrule
						\multicolumn{4}{c}{~}\\
						\multicolumn{4}{c}{~}\\
						\multicolumn{4}{c}{~}\\
						\multicolumn{4}{c}{~}\\
					\end{tabular}
		\end{center}}\end{minipage}
	}
	\caption{Ablation study of MRM using DeepLab-V2 \cite{chen2017deeplab} with ResNet-101 \cite{he2016deep} on the GTA $\rightarrow$ Cityscape benchmark with DACS \cite{tranheden2021dacs}.}
	\label{tab2}%
\end{table*}

\textbf{Rebuilder design.} The Rebuilder is designed to be lightweight, ensuring MRM’s auxiliary training introduces minimal overhead. Results in Tables~\ref{tab2}(a–b, d) show that increasing the embedding dimension and Transformer depth improves performance up to a point (best with dim=512, 2 blocks, $H^\prime=W^\prime=16$). Beyond this scale, performance drops due to instability when training deep Transformers atop pre-trained backbones. This suggests that moderate capacity stabilizes optimization while preserving auxiliary benefits, and future work could explore stronger yet stable Rebuilder designs.

\textbf{Masking vs. rebuilding.} To disentangle contributions, we separately test masking ``$(1-M^{s})\odot f^{t}$'' and rebuilding ``$M^{s}\odot f^{o}$''. As shown in Table~\ref{tab2}(c), masking alone slightly harms performance (–0.2 mIoU), indicating that feature-space masking causes irreversible semantic loss. Adding the rebuilding branch restores lost semantics and yields significant gains, confirming that reconstruction is essential to effective representation regularization.

\textbf{Training objective.} Table~\ref{tab2}(e) compares several reconstruction objectives. Pixel-level regression (as in MAE \cite{he2022masked}) underperforms (–0.3 mIoU) due to its low-level focus. Feature reconstruction using teacher \cite{bardes2024revisiting} or student features \cite{bardes2024revisiting} offers minor gains. In contrast, directly applying pixel-wise classification with cross-entropy loss yields the best result, emphasizing that auxiliary supervision should be task-aligned with segmentation.

\textbf{Trade-off.} Table~\ref{tab2}(f) shows that MRM’s performance is stable across a wide range of weighting factors. The best results are obtained with a trade-off of 1.0, while overly large weights (e.g., 10) slightly degrade performance, indicating that MRM requires little hyperparameter tuning.

\textbf{MRM on different domain.} In our main setup, MRM is applied to target-domain images for auxiliary training. Here, we examine its effect when applied to the source domain using DACS \cite{tranheden2021dacs} and DAFormer \cite{hoyer2022daformer}. As shown in Table~\ref{tab3}, performance gains arise only from target-domain MRM, while applying it to both domains brings no benefit and may even reduce mIOU. This result indicates that MRM primarily aids adaptation by refining target-domain features. In contrast, source-side reconstruction biases representations toward the source distribution, weakening domain alignment. Thus, effective masked modeling for UDA should emphasize target-domain reconstruction as an adaptive regularizer rather than a generic self-supervised task.

\begin{table}[htbp]
	\centering
	\small
	\setlength{\tabcolsep}{14pt}
	\begin{tabular}{ccc}
		\toprule
		MRM Domain & DACS \cite{tranheden2021dacs} & DAFormer \cite{hoyer2022daformer} \\
		\midrule
		–  & 52.1 & 68.3\\
		Source & 52.9 (+0.8) & 68.2 (-0.1) \\
		Target & 55.9 (+3.8) & 70.3 (+2.0)\\
		Source + Target & 55.2 (+3.1) & 69.0 (+0.7)\\ 
		\bottomrule
	\end{tabular}%
	\caption{MRM Performance across Source and Target Domains on the GTA$\rightarrow$Cityscapes Benchmark.}
	\label{tab3}%
\end{table}%

\textbf{Ablation on encoder and decoder training in MRM.}
Unlike contrastive-based auxiliary tasks \cite{xie2023sepico,chen2023pipa} that enhance only the encoder, MRM jointly optimizes both encoder and decoder, strengthening the entire segmentation pipeline. We conduct ablations using DACS \cite{tranheden2021dacs} and DAFormer \cite{hoyer2022daformer}, freezing either component during MRM training. As shown in Table~\ref{tab4}, performance drops notably when either part is fixed, confirming that MRM’s effectiveness relies on joint encoder–decoder optimization. This highlights a promising direction for designing auxiliary tasks that supervise the full network rather than isolated components.

\begin{table}[htbp]
	\centering
	\small
	\setlength{\tabcolsep}{10pt}
	\begin{tabular}{cc|cc}
		\toprule
		Encoder & Decoder & DACS \cite{tranheden2021dacs} & DAFormer \cite{hoyer2022daformer}  \\
		\midrule
		& & 52.1 & 68.3 \\
		\ding{52} & & 54.9 (+2.8) & 69.3 (+1.0) \\
		& \ding{52} & 54.2 (+2.1) & 69.1 (+0.8) \\
		\ding{52} & \ding{52} &  55.9 (+3.8) & 70.3 (+2.1)  \\
		\bottomrule
	\end{tabular}%
	\caption{Component-wise Performance Analysis of MRM Training on the GTA$\rightarrow$Cityscapes Benchmark.}
	\label{tab4}%
\end{table}%

\textbf{Further architectures.}
To assess generalization, we extend MRM to diverse encoder–decoder combinations and UDA methods, including ResNet-50/101 \cite{he2016deep}, MiT-B2/B3 \cite{xie2021segformer}, and decoders such as DeepLab-V3+ \cite{chen2018encoder}. We also evaluate with DACS \cite{tranheden2021dacs} and MIC \cite{hoyer2023mic} to cover different adaptation paradigms. As summarized in Table~\ref{tab5}, MRM consistently improves mIoU across all configurations, with gains persisting as model capacity increases—demonstrating its plug-and-play nature and scalability across architectures.

\begin{table}[htbp]
	\centering
	\small
	\setlength{\tabcolsep}{1.5pt}
	\begin{tabular}{cccccc}
		\toprule
		Encoder & Decoder & UDA & w/o MRM & w/ MRM \\
		\midrule
		RN50 \cite{he2016deep} & DLV2 \cite{chen2017deeplab} & DACS \cite{tranheden2021dacs} & 52.0 & 55.1 (+3.1)\\
		RN101 \cite{he2016deep} & DLV3+ \cite{chen2018encoder} & DACS \cite{tranheden2021dacs} & 54.7 & 59.3 (+4.6)\\
		RN101 \cite{he2016deep} & DLV2+ \cite{chen2017deeplab} & MIC \cite{hoyer2023mic} & 64.2 & 67.1 (+2.9)\\
		MiT-b2 \cite{xie2021segformer} & DAFH \cite{hoyer2022daformer} & DAFormer \cite{hoyer2022daformer} & 64.2 & 66.3 (+2.1)\\
		MiT-b3 \cite{xie2021segformer} & DAFH \cite{hoyer2022daformer} & MIC \cite{hoyer2023mic} & 73.6 & 75.8 (+2.2)\\
		\bottomrule
	\end{tabular}%
	\caption{Various UDA Methods with Guidance Training for mIOU Improvement on the GTA$\rightarrow$Cityscapes Benchmark.}
	\label{tab5}%
\end{table}%

\subsection{\textbf{Qualitative Analysis}}
\begin{figure*}[!t]
	\centering
	\includegraphics[width=6.8in]{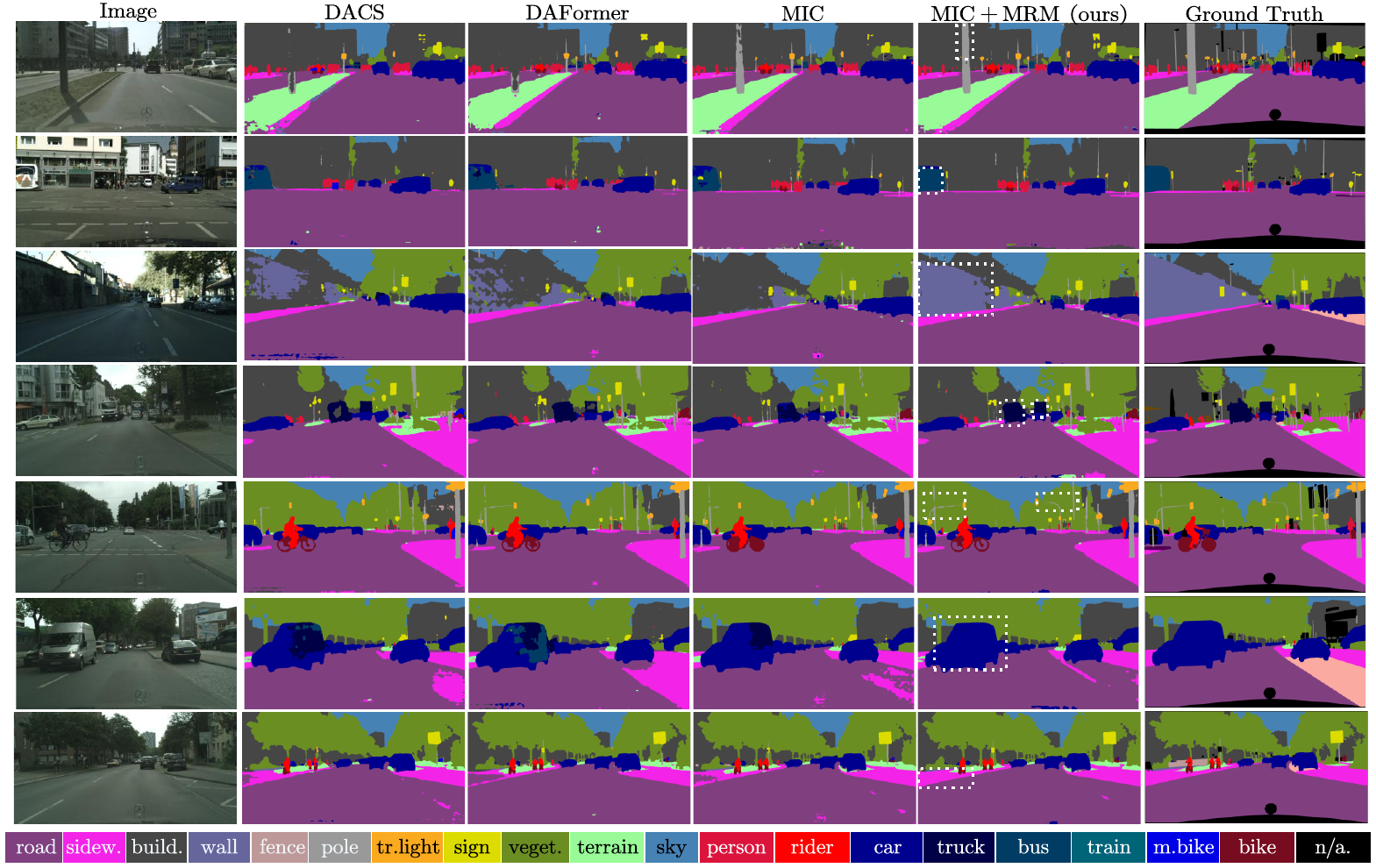}
	\caption{Qualitative comparison of MRM with previous methods on GTA $\rightarrow$ Cityscapes. To ensure a fair comparison and to demonstrate MRM's capability in contextual semantic consistency and long-range dependency modeling, we uniformly adopt the DeepLabv2 \cite{chen2017deeplab} with ResNet-101 \cite{he2016deep} architecture.}
	\label{fig5}
\end{figure*}

To ensure fair comparison, we evaluate all methods under a unified DeepLabV2–ResNet101 framework and visualize results of MRM integrated with MIC. As shown in Figure~\ref{fig5}, MRM produces cleaner and more coherent predictions than prior UDA approaches, revealing two key observations.

\textbf{Contextual semantic consistency.}
Existing UDA models often produce fragmented or contextually inconsistent regions—such as broken ``wall'' structures or mixed vehicle categories—due to missing target-domain supervision. MRM alleviates these issues by enforcing feature-level reconstruction, which strengthens local continuity and preserves structural semantics across class boundaries. This indicates that auxiliary reconstruction implicitly regularizes spatial context, promoting stable domain alignment and preventing label noise from propagating through pseudo-supervised training.

\textbf{Long-range dependency modeling.}
While MIC enhances local context, its CNN backbone restricts global reasoning. The Transformer-based Rebuilder enables MRM to capture long-range dependencies even within convolutional architectures, allowing better separation of visually similar but semantically distinct regions (e.g., ``wall'' vs. ``building''). Beyond accuracy gains, this suggests that MRM transfers a global awareness prior to the base network—encouraging more topology-consistent segmentation and improving robustness under severe appearance shifts. 
\label{sec:discussion}

\section{Conclusion}
We propose Masked Representation Modeling (MRM), a novel auxiliary task for unsupervised domain adaptive semantic segmentation. Unlike conventional masked image modeling, MRM operates directly in the latent space, avoiding architectural conflicts and task misalignment. By coupling representation reconstruction with the pixel-wise segmentation objective and involving the decoder in auxiliary supervision, MRM strengthens the entire segmentation pipeline without additional inference cost. Extensive experiments across multiple benchmarks and architectures validate its generalizability and effectiveness. Our analyses further reveal that representation reconstruction not only aids domain alignment but also enhances decoder regularization, improving the consistency of cross-domain representations. We hope this study provides insights into integrating task-aligned masked modeling as an effective auxiliary signal for domain adaptation.
\label{sec:acknowledgement}

\section*{Acknowledgements}
This work was supported by the National Natural Science Foundation of China (32572189), the Guangdong Basic and Applied Basic Research Foundation (2024A1515140111, 2026A1515011831, 2026A1515010362), the Guangzhou Science and Technology Project (2023B01J0011), the Shaoguan Science and Technology Project (230316116276286), the Foshan Science and Technology Project (2220001018608), the Zhuhai Science and Technology Project (2320004002668), and the Zhongshan Science and Technology Project (2024A1010).
{
    \small
    \bibliographystyle{ieeenat_fullname}
    \bibliography{mrm}
}

\clearpage
\setcounter{page}{1}
\maketitlesupplementary
\label{sec:rationale}

\begin{algorithm}[htbp]
	\small
	\caption{Training with Masked Representation Modeling (MRM).}
	\label{alg:mrm}
	\begin{algorithmic}[1]
		\REQUIRE Source images $\{x_s, y_s\}$, target images $\{x_t\}$, segmentation model $F = D \circ E$, Rebuilder $R$
		\FOR{each training step}
		\STATE $\#\quad \textit{Supervised loss on source domain}$
		\STATE $f_s \gets E(x_s)$
		\STATE $\hat{y}_s \gets D(f_s)$
		\STATE $\mathcal{L}_{\text{sup}} \gets \text{CrossEntropy}(\hat{y}_s, y_s)$
		\vspace{0.5em}
		
		\STATE $\#\quad \textit{Unsupervised loss (e.g., pseudo-labeling)}$
		\STATE $f_t \gets E(x_t)$
		\STATE $\hat{y}_t \gets D(f_t)$
		\STATE $\tilde{y}_t \gets \text{PseudoLabel}(\hat{y}_t)$
		\STATE $\mathcal{L}_{\text{uda}} \gets \text{CrossEntropy}(\hat{y}_t, \tilde{y}_t)$
		\vspace{0.5em}
		
		\STATE $\#\quad \textbf{Masked Representation Modeling (ours)}$
		\STATE $f_t^{\text{recon}} \gets R(f_t^{\text{mask}})$
		\STATE $\mathcal{L}_{\text{mrm}} \gets \text{CrossEntropy}(D(f_t^{\text{recon}}), \tilde{y}_t)$
		\vspace{0.5em}
		
		\STATE $\mathcal{L} \gets \mathcal{L}_{\text{sup}} + \mathcal{L}_{\text{uda}} + \lambda \cdot \mathcal{L}_{\text{mrm}}$
		\STATE Update model parameters via backpropagation
		\ENDFOR
	\end{algorithmic}
\end{algorithm}

\section{\textbf{Pseudo Code}}
The pseudo code in Algorithm~\ref{alg:mrm} outlines the integration of our proposed MRM into a typical domain adaptation training pipeline. MRM is designed to be modular and can be seamlessly incorporated into existing frameworks that combine supervised learning on source data and unsupervised learning on target data via pseudo-labeling \cite{lee2013pseudo}.

At each training step, conventional losses are computed: a supervised segmentation loss on the source domain, and an unsupervised loss on the target domain. Our method introduces an additional auxiliary branch, where masked representation are reconstructed and then regularized using the pseudo-label supervision. This not only enriches the representation but also enhances generalization across domains.

Importantly, MRM does not interfere with the original optimization objectives but complements them with minimal architectural changes. Its plug-and-play nature allows it to be effectively combined with a wide range of existing domain adaptation or semi-supervised segmentation methods.

\section{\textbf{Further Ablation Study}}

\textbf{Different Masking Locations in the Network.} MAE \cite{he2022masked} removes image patches and processes only the visible ones through the encoder, which makes it incompatible with semantic segmentation encoders that rely on full spatial inputs. To address this issue, MRM performs masking and reconstruction directly in the latent space, enabling compatibility with arbitrary backbone models as long as the input and output shapes of the Rebuilder are aligned. However, alternative designs are also possible. Inspired by BERT \cite{kenton2019bert}, one could insert masked tokens before the encoder by concatenating them with visible patches, allowing the encoder to maintain a standard forward pipeline. In this section, we explore the effectiveness of this design choice. The results are presented in Table~\ref{tab6}.

\begin{figure}[t]
	\centering
	\includegraphics[width=3.35in]{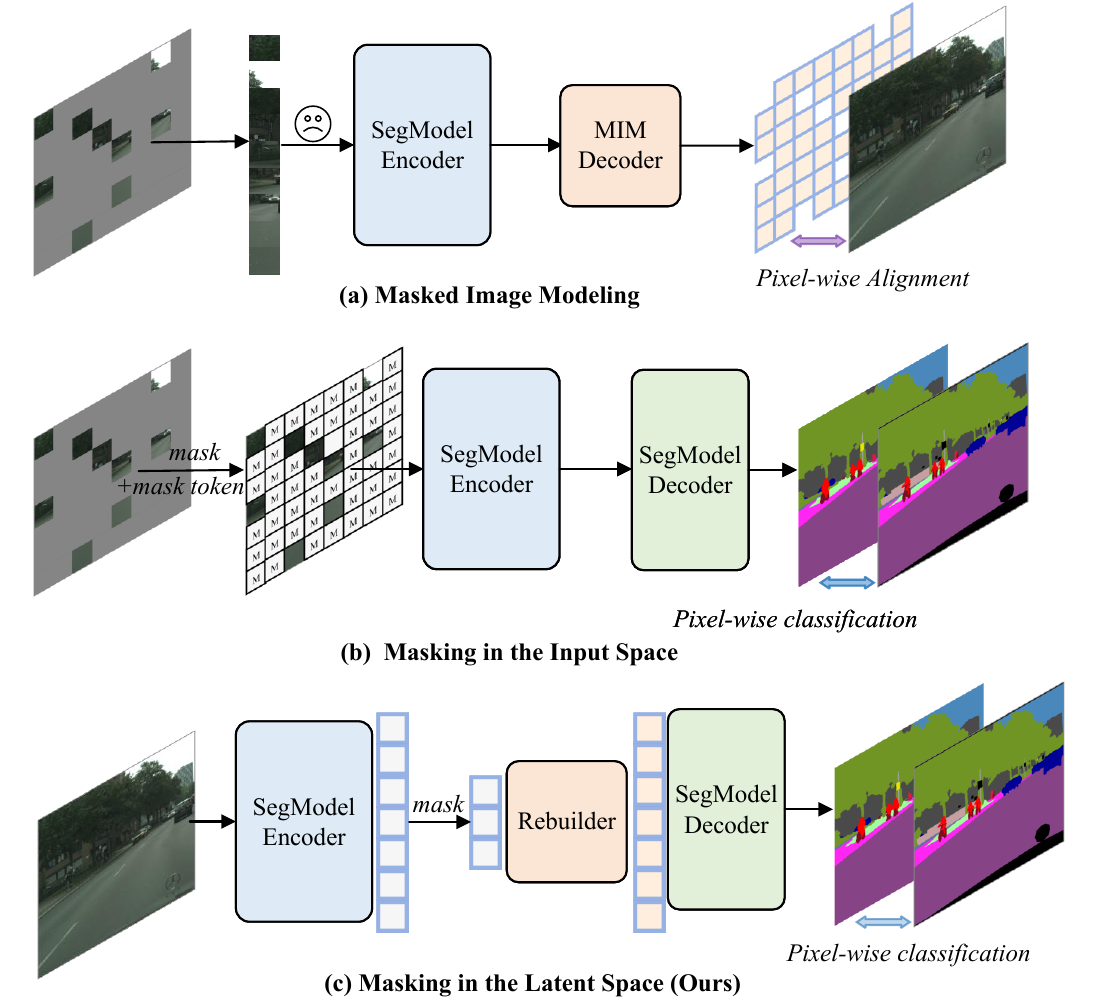}
	\caption{Different masking locations in the network.}
	\label{fig7}
\end{figure}

\begin{table}[htbp]
	\centering
	\small
	\setlength{\tabcolsep}{6pt}
	\begin{tabular}{ccc}
		\toprule
		Masking Position & mIOU & Training FLOPs (G)\\
		\midrule
		w/o MRM & 52.1 & 1.00$\times$ (182.34) \\
		\midrule
		Input Space & 54.0 (+1.9) & 2.00$\times$ (364.68)\\
		Latent Space (ours) & 55.9 (+3.8) & 1.13$\times$ (207.74)\\
		\bottomrule
	\end{tabular}%
	\caption{Ablation on masking locations using a DACS\cite{tranheden2021dacs} evaluated on the GTA $\rightarrow$ Cityscapes Benchmark.}
	\label{tab6}%
\end{table}%

Experiment demonstrate that applying masking in the input space can also lead to performance improvements. Notably, since the Rebuilder is removed during training, this indirectly validates the importance of using the main task loss as the optimization objective for the auxiliary task. However, masking in the latent space still yields greater performance gains compared to input-space masking.

To provide a more intuitive comparison, we also introduce Training Floating Point Operations (FLOPs) as a metric to evaluate the computational cost of the two masking strategies in MRM. Specifically, this FLOPs count reflects the cost incurred when processing the target image. For reference, the baseline method (i.e., DACS w/o MRM) requires 182.34G FLOPs. In the input-space design, after masking and inserting mask tokens, the resulting input is no longer identical to the original target image. As a result, the processed input must be passed through the full network again, effectively doubling the training FLOPs compared to the baseline.

In contrast, MRM maintains the original input (i.e., the target image), allowing it to reuse the encoder's representation already computed in the UDA pipeline. This avoids redundant computations and leads to substantial savings in training cost. The additional overhead comes only from passing these encoder features through the lightweight Rebuilder and Decoder modules, which are significantly more efficient than the encoder itself, resulting in only a marginal increase in FLOPs.

\section{\textbf{Discussion}}

\subsection{Limitations and Future Work}
While MRM demonstrates strong performance across diverse architectures and datasets, several limitations remain:

\begin{itemize}
	\item \textbf{Limited capacity of Rebuilder}: To maintain lightweight design, the Rebuilder employs only a small number of Transformer blocks. Although sufficient for current benchmarks, its ability to model complex feature dependencies may be limited in more diverse domains.
	
	\item \textbf{Training stability}: We observed that scaling up the Rebuilder or increasing the masking ratio can lead to unstable training. This reflects the difficulty of integrating auxiliary Transformer components with pre-trained segmentation models.
	
	\item \textbf{Task specificity}: MRM is specifically designed for pixel-wise classification tasks. Its direct extension to other dense prediction tasks such as depth estimation or panoptic segmentation requires further investigation.
\end{itemize}

In future work, we aim to explore more powerful and stable Rebuilder architectures, potentially leveraging pre-trained or distilled representations to improve semantic reconstruction, as well as incorporating diverse masking strategies to enhance robustness. Moreover, we plan to generalize the MRM paradigm to other structured prediction tasks, and to study its synergy with advanced pseudo label and consistency regularization techniques. Finally, we are interested in investigating the theoretical foundations of task-aligned auxiliary learning and its generalization benefits in domain adaptation.

\subsection{What makes the Rebuilder pluggable?} 

To enable MRM to be compatible with arbitrary architectures and simultaneously incorporate the decoder into auxiliary task training, we introduce a Rebuilder module between the encoder and decoder to perform reconstruction-based decoding during training. This results in an inference process of $D(R(E(x)))$. Notably, although this training strategy modifies the original pipeline, the Rebuilder can be seamlessly removed during inference, restoring the original inference flow $D(E(x))$.

The ability to remove the Rebuilder during inference without compromising performance is attributed to hybrid training strategy. Specifically, during training, the decoder is jointly exposed to both the reconstructed features $R(E(x))$ and the original features $E(x)$ derived from both source and target domain samples. This allows the decoder to learn from a joint distribution encompassing both types of features. Consequently, at inference time, even in the absence of reconstructed inputs, the decoder remains effective, as the distribution of original features is subsumed within the training-time joint distribution. This ensures that removing the Rebuilder has minimal to no adverse impact on the inference process.

\subsection{Difference from MIC}

Although both MIC~\cite{hoyer2023mic} and our Masked Representation Modeling use masking to improve domain-adaptive segmentation, they differ in where masking is applied and how the masked information is used.

MIC operates in the image space: masked images are passed through the network and the model is trained with a consistency objective between masked and unmasked predictions. In this sense, MIC is conceptually similar to a high-ratio CutOut-style augmentation, where large portions of the input are removed and the model learns robustness to severe input corruption.

In contrast, MRM performs masking in the latent representation space. Instead of masking pixels, we mask encoder features and reconstruct the missing representations from the visible ones, and the reconstructed features are optimized using the standard segmentation objective.

These mechanisms regularize the model at different levels. MIC primarily improves robustness to input-level corruption, while MRM enforces semantic consistency in feature space. As a result, the two approaches are conceptually distinct and often complementary in practice.

\subsection{Why does MRM work?} While prior work has demonstrated that masked image modeling (MIM) \cite{he2022masked, woo2023convnext, assran2023self} enhances the representation capacity of the backbone by training with masked inputs and strengthens contextual modeling—findings also supported by previous qualitative analyses—MRM operates differently. Unlike MIM, which masks input tokens in the image space, MRM performs occlusion and reconstruction directly in the latent space. As a result, the backbone processes the full input signal, rather than only the visible tokens as in MAE \cite{he2022masked} for vision or BERT \cite{kenton2019bert} for NLP. This distinction suggests that the effectiveness of MRM may stem from additional factors beyond those commonly attributed to MIM.

We observe that MRM exhibits notable similarities with feature-level mixup \cite{verma2019manifold}. Specifically, both employ a loss function aligned with the primary task during training, and their feature fusion strategies are remarkably analogous.

\begin{eqnarray}
	f^{r} = M^{s} \odot f^{o} +(1-M^{s}) \odot f^{t}
	\label{eq8}
\end{eqnarray}
\begin{eqnarray}
	f^{mix} = \lambda_{mix} \cdot f^{i} +(1-\lambda_{mix}) \cdot f^{j}
	\label{eq9}
\end{eqnarray}

Equation~(\ref{eq8}) specifies the MRM-based representation reconstruction operation, whereas Equation~(\ref{eq9}) represents feature-level mixup. Two primary distinctions can be observed: (i) MRM conducts fusion in a local region of the feature map, in contrast to the global fusion strategy used in standard feature-level mixup. Notably, feature-level mixup can also be extended to support local fusion. (ii) The sources of fusion differ—mixup leverages features from a different sample, while MRM generates a fusion target from the features of the same input. Given their structural similarity, MRM can be interpreted as a special case of feature-level mixup. This perspective enables us to analyze the effectiveness of MRM through the lens of feature-level mixup.

Beyond its structural resemblance to feature-level mixup, MRM can also be interpreted through the lens of the Information Bottleneck (IB) principle \cite{tishby2000information}. The IB framework posits that a good representation $Z$ should capture the minimal sufficient statistics of the input $X $ with respect to the target $Y$, by maximizing the mutual information $I(Z;Y)$ while minimizing $I(Z;X)$. In this context, MRM implicitly acts as an information bottleneck by enforcing localized stochastic fusion in the feature space, effectively injecting structured noise into the intermediate representations. Unlike mixup—which introduces global perturbations across samples—MRM perturbs features locally and within the same instance. This intra-sample stochasticity encourages the network to retain task-relevant, spatially consistent features while suppressing instance-specific redundancy. As such, MRM serves as a self-supervised regularizer that biases the learned representation toward compression and abstraction.

From an IB perspective, MRM reduces $I(Z;X)$ by removing local pixel-level details, while preserving or even enhancing $I(Z;Y)$, particularly when the reconstruction loss aligns with the downstream task. This enables MRM to compress input information in a manner that preserves its task-relevant aspects, improving generalization performance in downstream tasks.  

\end{document}